\documentclass{article} 
\usepackage{iclr2025_conference,times}

\usepackage{amsmath,amsfonts,bm}

\def\eqref#1{equation~\ref{#1}}

\def\1{\bm{1}}

\DeclareMathAlphabet{\mathsfit}{\encodingdefault}{\sfdefault}{m}{sl}
\SetMathAlphabet{\mathsfit}{bold}{\encodingdefault}{\sfdefault}{bx}{n}

\usepackage{hyperref}
\usepackage{url}

\title{Understanding Financial Reasoning in AI: A Multimodal Benchmark and Error Learning Approach}

\author{
Shuangyan Deng, Haizhou Peng, Jiachen Xu, Chouhou Liu,  
\\\textbf{Ciprian Doru Giurc\u{a}neanu, Jiamou Liu}\\
University of Auckland \\
Auckland, New Zealand \\
\texttt{\{sden118\}@aucklanduni.ac.nz}
}

\iclrfinalcopy 
\usepackage{graphicx}
\usepackage{xcolor}
\usepackage{colortbl} 
\usepackage{array}
\usepackage{booktabs}
\usepackage{caption}
\usepackage{tcolorbox} 
\usepackage{fontawesome} 
\usepackage{hyperref} 
\usepackage{enumitem}
\usepackage{titlesec}

\iclrfinalcopy 
\begin{document}

\maketitle

\begin{abstract}
Effective financial reasoning demands not only textual understanding but also the ability to interpret complex visual data such as charts, tables, and trend graphs. This paper introduces a new benchmark designed to evaluate how well AI models—especially large language and multimodal models—reason in finance-specific contexts. Covering 3,200 expert-level question-answer pairs across 15 core financial topics, the benchmark integrates both textual and visual modalities to reflect authentic analytical challenges in finance. To address limitations in current reasoning approaches, we propose an error-aware learning framework that leverages historical model mistakes and feedback to guide inference, without requiring fine-tuning. Our experiments across state-of-the-art models show that multimodal inputs significantly enhance performance and that incorporating error feedback leads to consistent and measurable improvements. The results highlight persistent challenges in visual understanding and mathematical logic, while also demonstrating the promise of self-reflective reasoning in financial AI systems. Our code and data can be found at \textcolor{blue}{\faGithub} \href{https://anonymous.4open.science/r/FinMR-1FDD/README.md}{\texttt{https://anonymous/FinMR/Code\&Data}}. The leaderboard can be found at \textcolor{blue}{\faTrophy} \href{https://anonymous.4open.science/w/FinMR-homepage-35EF/}{\texttt{https://anonymous/FinMR/Leaderboard}}.
\end{abstract}
\begin{figure}[h]
    \centering
    \includegraphics[width=1\textwidth]{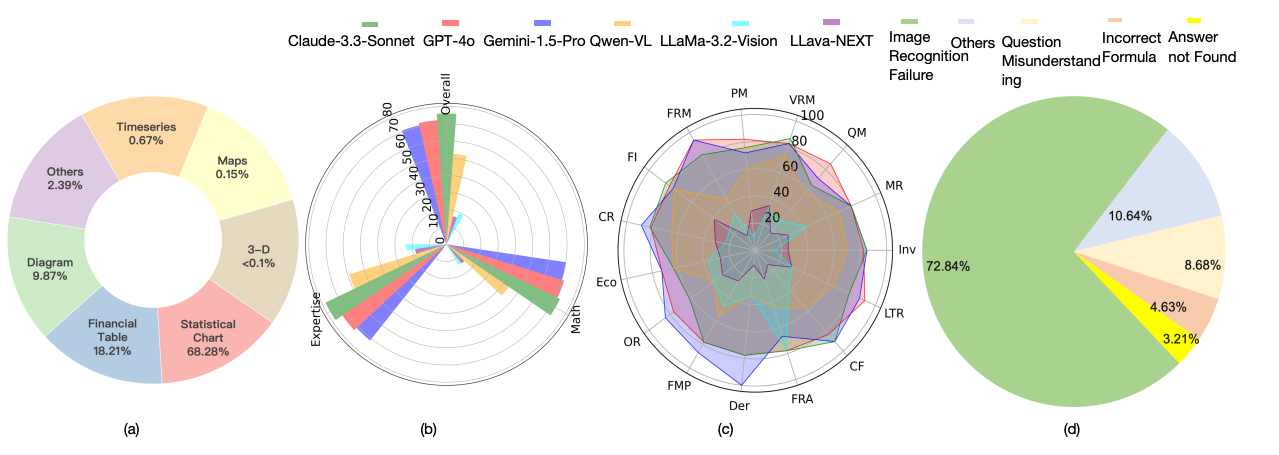}
    \caption{\textbf{FinMR} provides diverse visual data, as shown in panel (a). Evaluation of financial reasoning abilities of LLMs and MLLMs covers mathematical and expertise-based tasks (see panel (b)), and performance varies across 15 financial domain topics (see panel (c), the abbreviation list of topics provided in Table \ref{tab:topics}). The key errors shown in panel (d) categories include image recognition failures, incorrect formula application, question misunderstanding, and answer not found.}
    \label{fig:teaser}
\end{figure}

\vspace{-0.5 cm}
\section{Introduction}
\vspace{-0.2 cm}
\label{Sec: Intro}
Financial reasoning is a critical analysis process that involves leveraging expert-level knowledge to derive insights from diverse financial data and logically conclude a decision. Effective financial reasoning can lead to wise decisions that generate substantial monetary benefits or avoid costs amounting to billions of dollars \citep{jerven2013poor,mackenzie2008engine,chen2022FinQAdatasetnumerical}. However, analyzing and reasoning from financial data is inherently complex. Unlike general-domain reasoning tasks, financial reasoning  requires the integration of information from heterogeneous sources such as structured data (e.g., financial tables), semi-structured data (e.g., regulatory filings), and unstructured data (e.g., economic reports). These tasks demand advanced mathematical rigor, including multi-step calculations, statistical analysis, and domain-specific formulas for valuation, credit risk, and liquidity analysis. Additionally, financial reasoning relies on a deep understanding of complex concepts like portfolio optimization and risk modeling. 

Recent advancements in Large Language Models (LLMs), such as GPT-o1 and DeepSeek-R1, have significantly mitigated the challenges associated with reasoning over financial text data \citep{openaiIntroducingGPT4oMore2024, deepseekai2025deepseekr1incentivizingreasoningcapability}. However, many financial problems also involve visual data, such as stock price trends, financial tables, and statistical charts, which require integrating multimodal information for effective reasoning. Multimodal Large Language Models (MLLMs), capable of processing both textual and visual inputs, have demonstrated remarkable capabilities in addressing such complex tasks. These models have been evaluated on prominent multimodal benchmarks such as MMMU \citep{yueMMMUMassiveMultidiscipline2024} and Math Vista \citep{luMathVistaEvaluatingMathematical2024}, and documented in official technical reports \citep{openaiIntroducingGPT4oMore2024,geminiGemini15Unlocking2024,dubeyLlama3Herd2024,anthropicIntroducingComputerUse2024,li2024LLaVANeXTStrongerLLMs}.
Despite these advancements,  existing multimodal benchmarks (see Table \ref{tab:comparison_benchmarks}) primarily evaluate MLLM's general reasoning capabilities in open domains. The sole exception, FAMMA \citep{xue2024fammabenchmarkfinancialdomain}, is limited in scope, encompassing only 1758 examples across 8 topics. Key financial areas, such as risk management, valuation, and liquidity analysis, are inadequately represented, leaving the financial reasoning capabilities of MLLMs largely unexplored. To address this gap, we would like to answer the critical question:\textbf{\textit{ What are the multimodal reasoning capabilities and limitations of MLLMs in the financial domain?}}

To answer this question, first, we propose \textbf{FinMR}, a comprehensive benchmark specifically designed for evaluating the capabilities of MLLMs in financial reasoning tasks.  \textbf{FinMR} spans 15 diverse financial topics and includes 3,200 college-level question-answer pairs that combine textual and visual content.  These questions are carefully curated to encompass a wide range of visual data, including table images, stock price trends, and statistical distributions, as shown in Figure \ref{fig:teaser}(a). The benchmark is divided into 1,049 financial math questions, each requiring advanced mathematical skills such as calculus and statistics, and 2,151 financial expertise questions, which necessitate a deep understanding of domain-specific financial knowledge. Each example includes manually annotated expert explanations, facilitating both the evaluation of reasoning capabilities and detailed error analysis. To support model development and evaluation,  we split the dataset into an 80\% \textit{development} set (i.e., 2560 examples) and a 20\% \textit{test} set (640 examples).    

Second, we conduct a comprehensive evaluation of state-of-the-art LLMs and MLLMs on \textbf{FinMR}, identifying significant performance gaps across models. Notably, we test \textbf{\textit{LLMs with text input}}, including GPT-o1 \citep{openaiIntroductionOpenAIO12024}, Gemini-1.5-Pro \citep{geminiteam2024Geminifamilyhighly}, Claude-3.5-Sonnet \citep{anthropicIntroducingComputerUse2024}, Llama 3.2\citep{dubeyLlama3Herd2024}, Deepseek \citep{deepseekai2025deepseekr1incentivizingreasoningcapability}, and Qwen \citep{teamqwenQwen25LLMExtendingBoundary2024}, alongside \textbf{\textit{MLLMs with text \& image input}}, such as GPT-4o \citep{openaiIntroducingGPT4oMore2024}, Gemini-1.5-Pro \citep{geminiteam2024Geminifamilyhighly}, Claude-3.5-Sonnet \citep{anthropicIntroducingComputerUse2024}, Llama-3.2-Vision \citep{ai@meta2024Llama32Revolutionizing}, Qwen-VL-Plus \citep{baiQwenVLVersatileVisionlanguage2023}, LLaVa-NEXT \citep{li2024LLaVANeXTStrongerLLMs}. For evaluation, we adopt two methods, Chain-of-Thought (CoT) prompting \citep{wei2022Chainofthoughtpromptingelicits,li2024surveymultimodalcomposite} and our proposed {\em Error Feedback Learning (EFL)} method.  Inspired by prior studies that emphasize the value of error feedback in reasoning tasks \citep{yan2024surveymathematicalreasoning,wangExploringReasoningAbilities2024,lu2023surveydeeplearning}, we 
construct an {\em error database} using the development data. This database contains negative examples paired with AI-driven feedback and provides a foundation for the EFL method. EFL leverages our error database to retrieve similar negative examples and associated feedback.  This approach improves reasoning performance without the need for additional model training and supports effective retrieval-based reasoning.

Finally, we perform a detailed error analysis to identify key challenges in multimodal financial reasoning and provide guidance for future improvements. Our findings reveal that: (1) Multimodal inputs significantly enhance reasoning capabilities, though image recognition remains a critical bottleneck. Notably, Gemini-1.5-Pro (with text and image) achieves 82.06\% accuracy, while Gemini-1.5-Pro (with text and image caption) has only 61.37\% accuracy. This verifies that MLLM has higher advantages by direct image input; (2) EFL strategy comprehensively surpasses CoT, and the greatest improvement is 12.44\% of Qwen VL, indicating large models benefiting from error feedback; (3) Financial math reasoning remains particularly challenging, with models scoring approximately 10\% lower on math-related tasks compared to financial expertise tasks.  Figure \ref{fig:teaser}(d) highlights three primary error types in the reasoning process: image recognition failure, question misunderstanding, and incorrect formula application. Among them, image recognition is the most significant bottleneck, underscoring the need for advanced techniques in visual content understanding within the financial domain. 

The key contributions of this study are summarized as follows:
\begin{itemize}[leftmargin=*]
    \item  FinMR, the first comprehensive multimodal reasoning benchmark across 15 financial topics. The benchmark includes 3,200 question-answer pairs with explanations annotated manually, and each pair has text and diverse types of images, aiming at evaluating MLLM abilities in knowledge-intensive reasoning and analyzing their error in intermediate reasoning steps.

    \item EFL, a novel strategy to improve large models' reasoning abilities on \textbf{FinMR}. We construct a database of negative examples with AI-driven error feedback, which contributes to self-learning from previous mistakes and next iterate learning, evidencing the effectiveness of error feedback.

    \item Comprehensive experiments over mainstream LLMs and MLLMs which demonstrate that MLLMs consistently outperform standard LLMs. Our findings highlight the specific performance gap between these models, with Gemini-1.5-Pro emerging as the best-performing MLLM on \textbf{FinMR}. Notably, mathematical reasoning performance sees a significant improvement when direct image input is utilized.
    
\end{itemize}

\begin{table}[t]
    \centering
    \caption{Existing Reasoning Benchmarks versus FinMR}
    \label{tab:comparison_benchmarks}
    \resizebox{\textwidth}{!}{%
        \begin{tabular}{l|l|l|l|l|l|c|c|l}
            \hline
            \textbf{Benchmark} & \textbf{Domain} & \textbf{Modality} & \textbf{Level} & \textbf{Source} & \textbf{Number} & \textbf{Include Math?} & \textbf{Financial Expertise?} & \textbf{Solution Format} \\
            \hline
            MathVista~\cite{luMathVistaEvaluatingMathematical2024} & Open & Text \& Image & Elem. to College & Internet+Expert & 6141 & Yes & Few & Text \\
            MMMU~\cite{yueMMMUMassiveMultidiscipline2024} & Open & Text \& Image & College & Internet, Text-books, Lecture & 11500 & Yes & Few & Text \\
            MATH-V~\cite{wang2024Measuringmultimodalmathematical} & Math & Text \& Image & Elem., High School & Internet & 2252 & Yes & Few & Text \\
            FinQA~\cite{chen2022FinQAdatasetnumerical} & Finance & Text Only & College & Expert & 8281 & Yes & Yes & Math Program \\
            TAT-QA~\cite{zhu2021TATQAquestionanswering} & Finance & Text Only & College & Expert & 16552 & Yes & Yes & Text \\
            MultiHierrt~\cite{zhao2022MultiHierttNumericalreasoning} & Finance & Text Only & College & Expert & 10440 & Yes & Yes & Text \\
            DocMath-Eval~\cite{zhao2024DocMathevalEvaluatingmath} & Finance & Text Only & College & Internet+Expert & 5974 & Yes& Yes & Python Program \\
            FinanceMath~\cite{zhao2024FinanceMathKnowledgeintensivemath} & Financial Math & Text Only & College & Internet+Expert & 1200 & Yes & Yes & Python Program \\

             FAMMA~\cite{xue2024fammabenchmarkfinancialdomain} & Finance & Text \& Image & College & Textbook & 1758 & few & Yes & Text \\

            \hline
            \textbf{* FinMR (Ours)} & Finance & Text \& Image & College, Profession & Internet+Expert & 3700 & Yes & Yes & Text \\
            \hline
        \end{tabular}%
    }

\end{table}
\vspace{-0.5 cm}
\section{Related Works}
\vspace{-0.2 cm}
\subsection{Reasoning Benchmarks}
\vspace{-0.2 cm}
Early multimodal reasoning benchmarks, such as GeoQA \citep{chen2022geoqageometricquestionanswering} and GeoQA+ \citep{chen2022unigeounifyinggeometrylogical}, are narrow in scope, predominantly addressing plane geometry problems. More recent multimodal math reasoning datasets, such as MMMU \citep{yueMMMUMassiveMultidiscipline2024}, Math Vista \citep{luMathVistaEvaluatingMathematical2024}, and Math-Vision \citep{wang2024Measuringmultimodalmathematical},  broaden the subjects coverage and difficulty levels. However, these benchmarks focus on general mathematical reasoning and lack domain-specific content. Although FAMMA \citep{xue2024fammabenchmarkfinancialdomain} incorporates financial mathematical reasoning examples, it remains limited in scale, with only 1,758 examples across eight topics. Furthermore, FAMMA's scope does not fully capture the complexity of multimodal financial reasoning tasks. 

Several benchmarks evaluate textual reasoning in finance, focusing on financial statements and reports. For instance, TAT-QA \citep{zhu2021TATQAquestionanswering} combines tables and text for numerical reasoning, while FinQA \citep{chen2022FinQAdatasetnumerical} offers 8,281 expert-annotated QA pairs requiring math operations like addition and comparison. These tasks demand significant financial knowledge, making them more complex than typical QA. MultiHiertt \citep{zhao2022MultiHierttNumericalreasoning} and DocMath-Eval \citep{zhao2024DocMathevalEvaluatingmath} focus on tabular data for financial reasoning, with MultiHiertt incorporating hierarchical tables and unstructured text for complex tasks. FinanceMath \citep{zhao2024FinanceMathKnowledgeintensivemath} further combines text and tables with expert annotations. However, real-world financial data often includes diverse visuals like price charts, financial statement screenshots, and diagrams, which provide critical insights for reasoning. To address this, we introduce \textbf{FinMR}, a comprehensive multi-modal benchmark covering 15 financial topics and integrating seven visual data types, offering a robust foundation for evaluating multimodal reasoning in finance.

\subsection{Methods for Stimulating Inherent Reasoning Capability}

\textbf{Chain of Thought (CoT).} Reasoning capabilities in large models have traditionally been enhanced through pre-training and fine-tuning methods \citep{yan2024TabMedBERTtabularknowledge,liang2023inproceedingsliangUniMathFoundationalMultimodal2023addresssingapore,Shao2024DeepSeekMathPushingLimitsMathematicalReasoningOpenLanguageModels,liu2024Retrievalaugmentedmultimodalchainofthoughts}. While effective, these approaches often involve significant computational and time costs. In contrast, prompt-based methods such as  CoT prompting provide a more time-efficient and computationally cost-effective alternative \citep{wei2022Chainofthoughtpromptingelicits,kojimaLargelanguagemodels}. CoT enables models to articulate intermediate reasoning steps explicitly, which enhances their ability to process complex queries and arrive at accurate conclusions. This method has been integrated into QA systems, including financial reasoning tasks, to generate detailed reasoning steps before producing an answer \citep{chenFinQADatasetNumerical2021,zhu2021TATQAquestionanswering,zhao2022MultiHierttNumericalreasoning,zhao2024FinanceMathKnowledgeintensivemath,zhao2024DocMathevalEvaluatingmath}. Recent advancements have extended CoT  from textual reasoning  to multimodal domains, enabling models to process and reason across diverse modalities. Notable contributions in this area include the works of \citet{wang2024Measuringmultimodalmathematical}, \citet{yueMMMUMassiveMultidiscipline2024}, \citet{luMathVistaEvaluatingMathematical2024} and \citet{zhang2024Multimodalchainofthoughtreasoning}, which leverage CoT to enhance multimodal understanding and decision-making. These approaches enable models to process visual, textual, and other data types, allowing for more complex reasoning processes.  We will also adopt CoT prompting to evaluate MLLMs' financial reasoning capabilities on \textbf{FinMR}.

\noindent \textbf{Error Feedback.} Well-pre-trained LLMs and MLLMs possess an inherent learning capacity, which reduces their hallucination issues by leveraging external materials \citep{yu2024VisRAGVisionbasedretrievalaugmented,tan2024Retrievalmeetsreasoning,zhao2023Retrievingmultimodalinformation,liu2024Retrievalaugmentedmultimodalchainofthoughts} and simulating the given examples \citep{tsimpoukelliMultimodalFewshotLearning2021,chen2023seethinkconfirminteractive,wei2022emergentabilitieslargelanguage}.  This capacity has been further enhanced through the application of in-context learning, which has been extended to multimodal tasks, including complex reasoning \citep{liu2024Retrievalaugmentedmultimodalchainofthoughts,zhao2023KnowledgeMathKnowledgeintensivemath,zhang2024Multimodalchainofthoughtreasoning}. 
One promising approach within this paradigm is learning from error feedback, which involves using prior mistakes to improve reasoning performance. Several studies highlight the value of error feedback in enhancing model performance on multimodal mathematical reasoning tasks \citep{yan2024surveymathematicalreasoning, wangExploringReasoningAbilities2024,lu2023surveydeeplearning,sun2024reviewmultimodalexplainable}. Building on this foundation, our work proposes a novel EFL strategy to retrieve similar error feedback from an error database. This approach allows models to iteratively refine their reasoning capabilities by analyzing errors and leveraging corrective feedback. 

\vspace{-0.2 cm}
\section{The FinMR Benchmark}
\vspace{-0.2 cm}
\label{Sec:benchmark}
\subsection{Overview of FinMR}
\vspace{-0.2 cm}
We introduce the Financial Multimodal Reasoning (\textbf{FinMR}) benchmark, a curated resource designed to evaluate the financial reasoning capabilities of large models across diverse topics and multimodal contexts. \textbf{FinMR} encompasses 15 topics in finance, ranging from \textit{Investment} to \textit{Liquidity and Treasury Risk}, as detailed in Table~\ref{tab:topics}. The benchmark includes 3,200 high-quality QA pairs with explanations, split into 2,151 expertise-based QA pairs and 1,049 math QA pairs, as shown in Table~\ref{tab:Benchmark Statistics}. All questions in our benchmark are manually collected from financial exam papers at top universities and are available on the website\footnote{\url{https://www.studocu.com/en-nz}}, ensuring the dataset represents expert-level financial reasoning tasks. More details are presented in the Appendix \ref{A1: Comparisons with Existing Benchmarks}

\textbf{FinMR} evaluates three critical skills in MLLMs: (1) visual information understanding, (2) intensive domain-specific knowledge involvement in finance, and (3) reasoning. Unlike traditional benchmarks, \textbf{FinMR} presents significant challenges by requiring models to process and integrate diverse, heterogeneous image types, including financial tables, stock price trends, and statistical charts, alongside textual information. This benchmark extends beyond basic visual recognition to demand a sophisticated multimodal approach that combines advanced analytical capabilities with mathematical and financial expertise. 
\begin{table}
    \renewcommand{\arraystretch}{1} 
    \begin{center}
        \label{tab:topics}
        \begin{minipage}{0.48\linewidth} 
            \centering
            \caption{Financial Topic Distribution of FinMR.}
            \label{tab:topics}
            \resizebox{\linewidth}{!}{
                    \begin{tabular}{llr}
                        \hline
                        \textbf{Topic \& Abbreviation} & \textbf{Number} & \textbf{Ratio} \\ \hline
                        Investment (Inv)                & 371  & 11.6\% \\
                        Quantitative Methods (QM)       & 342  & 10.7\% \\
                        Valuation and Risk Models (VRM) & 318  & 9.9\%  \\
                        Financial Markets and Products (FMP) & 297  & 9.3\%  \\
                        Financial Reporting and Analysis (FRA) & 264  & 8.3\%  \\
                        Portfolio Management (PM)       & 258  & 8.1\%  \\
                        Fixed Income (FI)               & 251  & 7.8\%  \\
                        Credit Risk (CR)                & 170  & 5.3\%  \\
                        Foundation of Risk Management (FRM) & 169  & 5.3\%  \\
                        Economics (Eco)                 & 156  & 4.9\%  \\
                        Operational Risk (OR)           & 131  & 4.1\%  \\
                        Derivatives (Der)               & 126  & 3.9\%  \\
                        Market Risk (MR)                & 121  & 3.8\%  \\
                        Corporate Finance (CF)          & 119  & 3.7\%  \\
                        Liquidity and Treasury Risk (LTT) & 107  & 3.3\%  \\ \hline
                    \end{tabular}
            }
        \end{minipage}
        \hfill
        \begin{minipage}{0.48\linewidth} 
            \centering
            \caption{FinMR Benchmark Statistics.}
            \label{tab:Benchmark Statistics}
            \resizebox{\linewidth}{!}{
                \begin{tabular}{lcc}
                    \hline
                    \textbf{Statistics} & \textbf{Number} & \textbf{Ratio} \\ 
                    \hline
                    \textbf{Total Questions} & 3200 & 100\% \\ 
                    * Test & 640 & 20\% \\ 
                    * Develop & 2560 & 80\% \\ 
                    \hline
                    \textbf{Total Images} & 3764 & 100\% \\ 
                    * QA with Single Image/\# of Images & 2643/2643 & 70\% \\ 
                    * QA with Multiple Images/\# of Images & 557/1118 & 30\% \\
                    \hline
                    \textbf{Reasoning Type} & & \\ 
                    * Expertise Reasoning QA & 2151 & - \\ 
                    * Math Reasoning QA & 1049 & - \\ 
                    \hline
                    \textbf{Average Length} & & \\ 
                    * Context & 327.74 & - \\ 
                    * Question & 33.97 & - \\ 
                    * Explanation & 63.24 & - \\ 
                    \hline
                \end{tabular}
            }
        \end{minipage}
    \end{center}
\end{table}

\vspace{-0.2 cm}
\subsection{Data Collection and Compliance.} 
The data collection process for \textbf{FinMR} was conducted in two stages. In the first stage, we compiled a collection of financial exam papers from both college-level courses and professional certification programs. In particular, we focused on exam papers from business schools that officially collaborate with Chartered Financial Analyst (CFA) and Financial Risk Management (FRM), which are internationally recognized institutions that provide exams for financial certificates. 
The curricula of these institutions are often integrated into university course designs. Consequently, final exam papers from these programs represent a valuable resource for developing expert-level reasoning tasks. We extracted all QA pairs from past final exam papers from these business schools. For exam questions available in PDF format, we utilized the Mathpix API \citep{wang2024Measuringmultimodalmathematical} to extract textual content, mathematical formulas, and images. To maintain consistency in dataset formatting, we extracted only images present in the questions while excluding those from options and explanations. In the second stage, we enlisted two PhD students specializing in Finance, both of whom have passed the CFA and FRM exams. These experts manually verified the correctness of the explanations and filtered out QA pairs that lacked correct answers or high-quality explanations. This rigorous verification process ensured that the dataset comprised QS pairs with detailed expert-validated explanations. 

\subsection{Data Quality Assurance.}

We employed a three-stage data curation process with six annotators (four master’s and two PhD students from Computer Science and Finance).  In the first stage, we categorized questions as expertise-based or math reasoning using GPT-4o and assigned reference topics. Four annotators manually verified the alignment between questions, images, options, answers, and explanations, correcting extraction errors. Image clarity was enhanced using Image Generator Pro, and non-English or incomplete explanations were removed, resulting in 4,470 QA pairs (30\% math-focused financial QA, 70\% financial expertise QA).  In the second stage, two PhD students reviewed the validity, completeness, and clarity of explanations. They expanded reasoning steps and addressed grammatical and stylistic issues to ensure high-quality financial reasoning. After this refinement, the dataset was reduced to 3,200 high-quality QA pairs. In the final stage, each question was labeled with relevant metadata, including question ID, source topics, and question type. The dataset was then split by topic: 80\% (2,560 samples) for development and 20\% (640 samples) for testing.

\section{Evaluation Framework}

This section outlines our evaluation framework for assessing the financial reasoning capabilities of large models using \textbf{FinMR}. Specifically, we discuss the error feedback database construction, the evaluation process for large models, and the prompting methods used in our experiments. 

\subsection{Error Database Construction}\label{sec:EDC}
\begin{figure}
    \centering
    \includegraphics[width=0.8\textwidth]{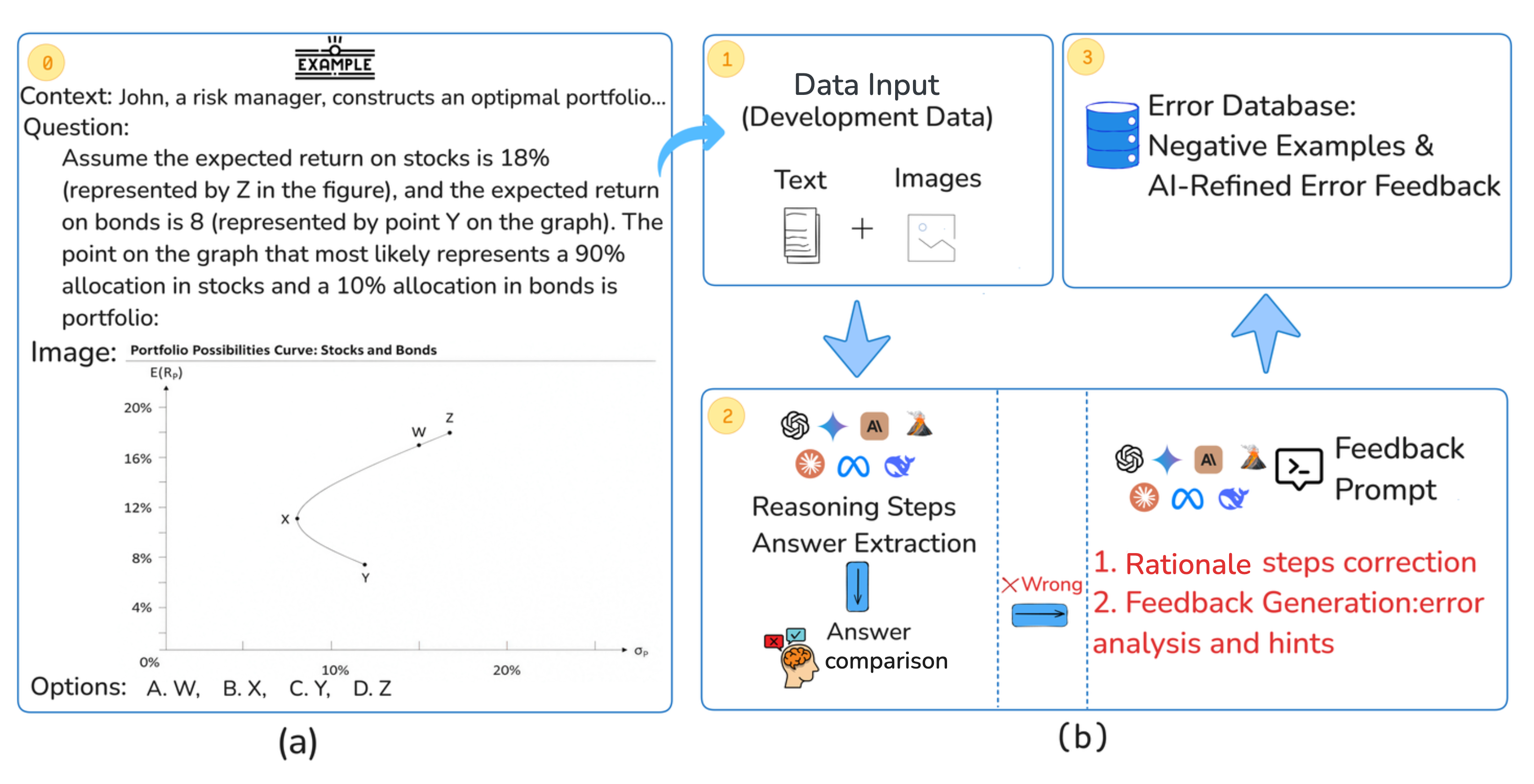}
    \caption{Panel (a) provides a typical example of FinMR. Panel (b) exhibits three stages of error database construction.  In the second stage, Large models leverage the annotated explanations to generate correct reasoning steps and hits.}
    \label{fig: Error Database}
\end{figure} 
A key component of our framework is the construction of an error feedback database, which is integral to the Error Feedback Learning (EFL) method. This database enables systematic analysis of model errors and facilitates iterative refinement of reasoning capabilities. 
The construction process, shown in Figure \ref{fig: Error Database}, involves three stages: {\em data input}, {\em feedback generation}, and {\em storage}. 

In the data input stage, we input context, questions, images, and options from the \textbf{FinMR} {\em development dataset} into the evaluated models. To accommodate  LLMs that cannot directly process images, we use GPT-4o  to convert images into textual descriptions, whereas MLLMs directly process the visual data alongside textual content. This approach follows established methodologies, such as those proposed in \citet{wang2024Measuringmultimodalmathematical}. 
In the feedback generation stage, models generate step-by-step reasoning and derive final answers. We then compare these answers against ground truth. For incorrect responses, we employ a feedback prompt (depicted in Figure \ref{fig:feedback_prompt} ) to guide the model in refining its reasoning steps. This process is further supported by manually annotated explanations. The generated feedback includes error analysis which elaborates on the specific reasoning flaws, and actionable guidance for addressing similar problems in the future. Finally, in storage stage, we save all relevant data, including the input examples, refined error feedback, and metadata (e.g., question ID, model information)in an external database.

\begin{minipage}{0.53\textwidth}
    \centering
    \footnotesize
    \begin{tcolorbox}
        \textbf{[System Input]:}

        You are a financial expert. You will be given questions and options, possibly with context information and images. Also, you will be given wrong reasoning steps and correct reasoning hints. You are supposed to give feedback in Markdown format. The feedback includes:\\
            1. Refine correct reasoning steps with a given explanation.\\
            2. Compare the correct reasoning steps and the model's wrong reasoning steps and highlight the difference.\\
            3. Summarize the hint for future similar questions.\\

        \textbf{[User Input]:}

        Context: \{context\}; Images in Context: \{images\}; Question: \{question\}; Images in Question: \{images\}; Options: \{options\}; Wrong Answer: \{model answer\}; Wrong Reasoning Steps: \{model reasoning steps\}; Correct Answer: \{answer\}; Explanation: \{explanation\}. 
    \end{tcolorbox}
    \captionof{figure}{Feedback Prompt Template}
    \label{fig:feedback_prompt}
\end{minipage}
\hfill
\begin{minipage}{0.47\textwidth}
    \centering
    \footnotesize
    \begin{tcolorbox}
        \textbf{[System Input]:}

        You are a financial expert and are supposed to answer the given questions with options, context information, and images. Also, You will be given previous learning documents, including questions and options, possibly with context information and images. Please answer the current question. The output reasoning steps are in Markdown format. Finally, you must put the correct option (A, B, C, or D) in \textbf{[ ]}. 

        e.g. Therefore, the correct option is \textbf{[B]}.\\

        \textbf{[User Input]:}
        
        Retrieved Example: \{example\}; Context: \{context\}; Images in Context: \{images\}; Question: \{question\}; Images in Question: \{images\}; Options: \{options\} \\

        Let's think step by step to answer the given question.

    \end{tcolorbox}
    \captionof{figure}{EFL Prompt Template}
    \label{fig:EFL Prompt Template}
\end{minipage}
\vspace{-0.5 cm}
\subsection{Evaluation Process}
\begin{figure}
    \centering
    \includegraphics[width=0.8\textwidth]{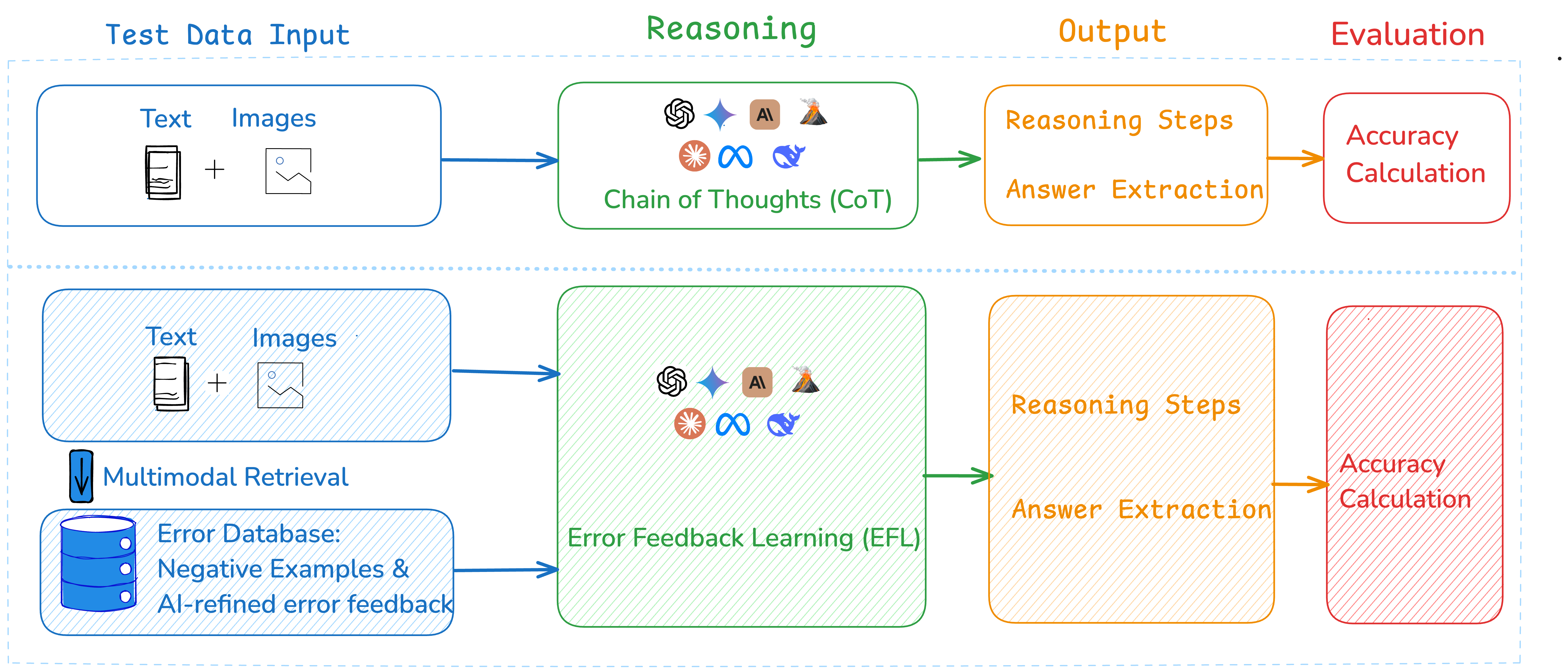}
    \caption{Four stages of the evaluation process. For LLMs with no visual ability, we leverage GPT-4o to generate image captions to support reasoning tasks. The process adopts two methods,  CoT and EFL. The latter retrieves the most similar (i.e., top-1 semantic similarity) negative examples and error feedback for learning.}
    \label{fig: Evaluation Process}
\end{figure}

The evaluation process consists of four stages: \textit{test data input, reasoning, output, and evaluation}, as displayed in Figure \ref{fig: Evaluation Process}. The test data input stage follows the same methodology as the \textit{data input stage} described in the construction of the error feedback database, including how we accommodate both LLMs and MLLMs (see Section~\ref{sec:EDC}), with the key distinction that the evaluation is performed using the test dataset.

The reasoning stage employs two distinct methods: CoT and EFL. CoT prompting involves guiding models to generate step-by-step reasoning through simple instructions, such as \textit{``Let's think step by step''} followed by the user input. In EFL, the model retrieves the most similar negative example and its error feedback from the previously constructed error database. The EFL prompt, presented in Figure \ref{fig:EFL Prompt Template}, incorporates this feedback into the reasoning process. The goal is to allow the model to learn from prior mistakes and refine their reasoning steps. This iterative retrieval mechanism is a novel contribution. For both reasoning methods, we clarify the format of the reasoning outputs using markdown, following practices outlined in  \citep{zhao2024DocMathevalEvaluatingmath,wang2024Measuringmultimodalmathematical}. 

In the output stage, we adopt the answer extraction pipeline inspired by \citep{chen2024M$^3$CoTnovelbenchmark,zhao2024FinanceMathKnowledgeintensivemath}. If the final answer is encapsulated in double square brackets (e.g., \texttt{[A]}), it is directly identified as the model's response. If no such format is detected, the output is categorized as \textit{``Answer not Found''}, which is regarded as an incorrect response. In the evaluation stage, the extracted answers are compared against the ground truth. The accuracy ratio is computed as the proportion of correct responses to the total number of questions. This is the primary metric for our evaluation.

\vspace{-0.2 cm}
\section{Results}
\vspace{-0.2 cm}
\subsection{LLM, MLLM, and Experiment Setup}
We evaluate the following LLMs on \textbf{FinMR}: 
\begin{itemize}[leftmargin=*]
    \item \textbf{Closed-source:} GPT-o1 \citep{openaiIntroductionOpenAIO12024}, Gemini 1.5 Pro \citep{geminiteam2024Geminifamilyhighly}, Claude-3.5-Sonnet \citep{anthropicIntroducingComputerUse2024}, Deepseek \citep{deepseekai2025deepseekr1incentivizingreasoningcapability};
    \item \textbf{Open-source:} Llama 3.2 \citep{dubeyLlama3Herd2024}, Qwen \citep{teamqwenQwen25LLMExtendingBoundary2024}.
\end{itemize}
%
We also evaluate the following closed-source and open-source MLLMs on \textbf{FinMR}:
\begin{itemize}[leftmargin=*]
    \item \textbf{Closed-source:} GPT-4o \citep{openaiIntroducingGPT4oMore2024}, Gemini-1.5-Pro \citep{geminiteam2024Geminifamilyhighly}, Claude-3.5-Sonnet \citep{anthropicIntroducingComputerUse2024}; 
    \item \textbf{Open-source:} Llama-3.2-Vision \citep{ai@meta2024Llama32Revolutionizing}, Qwen-VL-Plus \citep{baiQwenVLVersatileVisionlanguage2023}, LLaVa-NEXT \citep{li2024LLaVANeXTStrongerLLMs}.
\end{itemize}

All experiments on open-source models were conducted using A100 GPUs, while experiments on closed-source models were performed using 4090 GPUs. Additionally, we used LangSmith to trace all the experiments and set the temperature of large models to $0.7$.

\subsection{Evaluation Results}
\vspace{-0.3 cm}
\begin{table}[t]
    \centering
    \caption{Reasoning Performance Comparison of LLMs and MLLMs on \textbf{FinMR}. We highlight the best model's performance in green (LLMs) and blue (MLLMs).}
    \label{tab:Evaluation Result 1}
    \resizebox{\textwidth}{!}{ 
        \begin{tabular}{lllll|llllllllllllllll}
        \hline
        \\
        \multicolumn{1}{l}{}&{\textbf{Model}} & \textbf{Method} & \textbf{Overall} & \textbf{Math} & \textbf{Expertise}&
         \textbf{Inv}& 
          \textbf{QM} &
          \textbf{VRM}& 
          \textbf{FMP} &
          \textbf{FRA}& 
          \textbf{PM}& 
          \textbf{FI}& 
          \textbf{FRM}& 
          \textbf{CR}& 
          \textbf{Eco} &
          \textbf{OR} &
          \textbf{Der}& 
          \textbf{MR}& 
          \textbf{CF}& 
          \textbf{LTR}\\ 
        \\ \hline
        \multicolumn{20}{c}{\textbf{\textcolor{green!50!black}{Textual Modality: Text + Image Caption}}} \\
        \textbf{Closed-source} 
        & Claude-3.5-Sonnet& CoT & 53.91 & 42.91 & 61.83 & 52.50 & 30.00 & 65.38 & 52.17 & 22.22 & 48.10 & 60.24 & 37.50 & 52.83 & 18.18 & 30.77 & 22.22 & 65.00 & 47.62 & 56.52 \\
        & Claude-3.5-Sonnet& EFL & \textcolor{green!50!black}{64.84} & \textcolor{green!50!black}{55.97} & \cellcolor{green!10}\textcolor{green!50!black}{71.24} & \textcolor{green!40!black}{65.00} & \textcolor{green!40!black}{50.00} & \textcolor{green!40!black}{78.85} & \textcolor{green!40!black}{65.22} & \textcolor{green!40!black}{22.22} & \textcolor{green!40!black}{60.76} & \cellcolor{green!20}\textcolor{green!40!black}{71.08} & \textcolor{green!40!black}{62.50} & \textcolor{green!40!black}{60.38} & \textcolor{green!40!black}{45.45} & \textcolor{green!40!black}{46.15} & \textcolor{green!40!black}{33.33} & \textcolor{green!40!black}{75.00} & \textcolor{green!40!black}{61.90} & \textcolor{green!40!black}{56.52}\\
        & GPT-o1& CoT & 46.56 & 33.96 & 55.65 & 55.00 & 35.71 & 59.62 & 56.52 & 33.33 & 37.97 & 53.01 & 18.75 & 45.28 & 18.18 & 38.46 & 66.67 & 65.67 & 40.48 & 30.43 \\
         &GPT-o1& EFL & \textcolor{green!50!black}{60.31} & \textcolor{green!50!black}{49.25} & \textcolor{green!50!black}{68.28}& \textcolor{green!40!black}{67.50} & \textcolor{green!40!black}{50.00} & \textcolor{green!40!black}{69.23} & \textcolor{green!40!black}{65.22} & \textcolor{green!40!black}{55.56} & \textcolor{green!40!black}{49.37} & \textcolor{green!40!black}{68.67} & \textcolor{green!40!black}{31.25} & \textcolor{green!40!black}{61.32} & \textcolor{green!40!black}{36.36} & \textcolor{green!40!black}{46.15} & \cellcolor{green!20}\textcolor{green!40!black}{77.78} & \textcolor{green!40!black}{65.83} & \textcolor{green!40!black}{54.76} & \textcolor{green!40!black}{47.83} \\
        & Gemini-1.5-Pro& CoT & 47.81 & 34.70 & 57.26 & 47.50 & 28.57 & 57.69 & 60.87 & 33.33 & 36.71 & 53.01 & 6.25 & 49.06 & 54.55 & 38.46 & 55.56 & 61.67 & 35.71 & 21.74 \\
        & Gemini-1.5-Pro& EFL & \textcolor{green!50!black}{61.37} & \textcolor{green!50!black}{51.49} & \textcolor{green!50!black}{68.45} & \textcolor{green!40!black}{60.00} & \textcolor{green!40!black}{50.00} & \textcolor{green!40!black}{65.38} & \textcolor{green!40!black}{69.57} & \cellcolor{green!20}\textcolor{green!40!black}{77.78} & \textcolor{green!40!black}{46.84} & \textcolor{green!40!black}{70.59} & \textcolor{green!40!black}{18.75} & \textcolor{green!40!black}{65.09} & \cellcolor{green!20}\textcolor{green!40!black}{72.73} & \cellcolor{green!20}\textcolor{green!40!black}{61.54} & \textcolor{green!40!black}{55.56} & \cellcolor{green!20}\textcolor{green!40!black}{74.17} & \textcolor{green!40!black}{42.86} & \textcolor{green!40!black}{39.13} \\
        & DeepSeek-R1 & CoT & 61.25 & 66.42 & 57.53 & 65.00 & 64.29 & 61.54 & 78.26 & 55.56 & 53.16 & 59.04 & 56.25 & 61.32 & 54.55 & 46.15 & 66.67 & 72.50 & 83.33 & 73.91  \\
        & DeepSeek-R1 & EFL & \cellcolor{green!30}\textcolor{green!50!black}{71.88} & \cellcolor{green!10}\textcolor{green!50!black}{79.10} & \textcolor{green!50!black}{66.67}  & \cellcolor{green!20}\textcolor{green!40!black}{80.00} & \cellcolor{green!20}\textcolor{green!40!black}{78.57} & \cellcolor{green!20}\textcolor{green!40!black}{76.92} & \cellcolor{green!20}\textcolor{green!40!black}{91.30} & \textcolor{green!40!black}{66.67} & \cellcolor{green!20}\textcolor{green!40!black}{65.82} & \textcolor{green!40!black}{63.86} & \cellcolor{green!20}\textcolor{green!40!black}{68.75} & \textcolor{green!40!black}{70.75} & \cellcolor{green!20}\textcolor{green!40!black}{72.73} & \textcolor{green!40!black}{46.15} & \textcolor{green!40!black}{66.67} & \textcolor{green!40!black}{72.50} & \textcolor{green!40!black}{83.33} & \textcolor{green!40!black}{73.91}\\ \hline
        \textbf{Open-source}
        & LLaMa 3.2 & CoT & 27.34 & 23.51 & 30.11 & 27.50 & 21.43 & 23.08 & 39.13 & 0.00 & 25.32 & 36.14 & 18.75 & 23.58 & 9.09 & 7.69 & 11.11 & 32.50 & 35.71 & 21.74\\
        & LLaMa 3.2 & EFL & \textcolor{green!50!black}{36.09} & \textcolor{green!50!black}{32.09} & \textcolor{green!50!black}{38.98} & \textcolor{green!40!black}{40.00} & \textcolor{green!40!black}{21.43} & \textcolor{green!40!black}{28.85} & \textcolor{green!40!black}{52.17} & \textcolor{green!40!black}{33.33} & \textcolor{green!40!black}{35.44} & \textcolor{green!40!black}{46.99} & \textcolor{green!40!black}{18.75} & \textcolor{green!40!black}{26.42} & \textcolor{green!40!black}{9.09} & \textcolor{green!40!black}{7.69} & \textcolor{green!40!black}{44.44} & \textcolor{green!40!black}{43.33} & \textcolor{green!40!black}{42.86} & \textcolor{green!40!black}{34.78}  \\
        & Qwen & CoT & 55.62 & 52.99 & 57.53 & 55.00 & 35.71 & 51.92 & 52.17 & 33.33 & 50.63 & 60.24 & 31.25 & 62.26 & 45.45 & 30.77 & 55.56 & 57.50 & 71.43 & 56.52 \\
        & Qwen & EFL & \textcolor{green!50!black}{67.97} & \textcolor{green!50!black}{68.28} & \textcolor{green!50!black}{67.74} & \textcolor{green!40!black}{75.00} & \textcolor{green!40!black}{35.71} & \textcolor{green!40!black}{67.31} & \textcolor{green!40!black}{69.57} & \textcolor{green!40!black}{55.56} & \textcolor{green!40!black}{63.29} & \textcolor{green!40!black}{66.27} & \textcolor{green!40!black}{56.25} & \cellcolor{green!20}\textcolor{green!40!black}{76.42} & \cellcolor{green!20}\textcolor{green!40!black}{72.73} & \textcolor{green!40!black}{46.15} & \textcolor{green!40!black}{55.56} & \textcolor{green!40!black}{65.00} & \cellcolor{green!20}\textcolor{green!40!black}{80.95} & \cellcolor{green!20}\textcolor{green!40!black}{78.26} \\ \hline
        \multicolumn{20}{c}{\textbf{\textcolor{blue!50!black}{Multimodality: Text + Image}}} \\
        \textbf{Closed-source} 
        & GPT-4o& CoT & 72.19 & 71.43 & 73.17 & 65.00 & 71.43 & 71.15 & 73.91 & 66.67 & 73.42 & 74.70 & 81.25 & 71.70 & 63.64 & 38.46 & 77.78 & 73.33 & 76.19 & 78.26 \\
        & GPT-4o& EFL & \textcolor{blue!50!black}{81.72} & \textcolor{blue!50!black}{83.08} & \textcolor{blue!50!black}{81.03}  & \textcolor{blue!40!black}{82.50} & \cellcolor{blue!30}\textcolor{blue!40!black}{85.71} & \textcolor{blue!40!black}{82.69} & \textcolor{blue!40!black}{78.26} & \cellcolor{blue!30}\textcolor{blue!40!black}{77.78} & \cellcolor{blue!30}\textcolor{blue!40!black}{82.28} & \textcolor{blue!40!black}{80.72} & \cellcolor{blue!30}\textcolor{blue!40!black}{93.75} & \textcolor{blue!40!black}{81.13} & \cellcolor{blue!30}\textcolor{blue!40!black}{72.73} & \cellcolor{blue!30}\textcolor{blue!40!black}{76.92} & \textcolor{blue!40!black}{77.78} & \cellcolor{blue!30}\textcolor{blue!40!black}{80.00} & \textcolor{blue!40!black}{83.33} & \cellcolor{blue!30}\textcolor{blue!40!black}{91.30} \\
        & Gemini-1.5-Pro& CoT & 70.83 & 70.26 & 71.24 & 82.50 & 50.00 & 69.23 & 69.57 & 55.56 & 64.56 & 64.29 & 81.25 & 74.53 & 63.64 & 53.85 & 88.89 & 72.50 & 83.33 & 69.57\\
        & Gemini-1.5-Pro& EFL & \cellcolor{blue!30}\textcolor{blue!50!black}{82.06} & \cellcolor{blue!30}\textcolor{blue!50!black}{83.27} & \cellcolor{blue!30}\textcolor{blue!50!black}{81.18} & \textcolor{blue!40!black}{85.00} & \textcolor{blue!40!black}{71.43} & \textcolor{blue!40!black}{82.69} & \cellcolor{blue!30}\textcolor{blue!40!black}{86.96} & \textcolor{blue!40!black}{66.67} & \textcolor{blue!40!black}{72.15} & \textcolor{blue!40!black}{77.38} & \textcolor{blue!40!black}{93.75} & \cellcolor{blue!30}\textcolor{blue!40!black}{88.68} & \cellcolor{blue!30}\textcolor{blue!40!black}{72.73} & \textcolor{blue!40!black}{84.62} & \cellcolor{blue!30}\textcolor{blue!40!black}{100.00} & \cellcolor{blue!30}\textcolor{blue!40!black}{80.00} & \cellcolor{blue!30}\textcolor{blue!40!black}{90.48} & \textcolor{blue!40!black}{86.96}  \\
        & Claude-3.5-Sonnet& CoT & 75.94 & 73.13 & 77.96 & 75.00 & 57.14 & 80.77 & 69.57 & 77.78 & 68.35 & 79.52 & 75.00 & 76.42 & 63.64 & 53.85 & 66.67 & 76.67 & 88.10 & 73.91  \\
        & Claude-3.5-Sonnet& EFL & \textcolor{blue!50!black}{80.78} & \textcolor{blue!50!black}{80.22} & \cellcolor{blue!30}\textcolor{blue!50!black}{81.18} & \cellcolor{blue!30}\textcolor{blue!40!black}{85.00} & \textcolor{blue!40!black}{64.29} & \cellcolor{blue!30}\textcolor{blue!40!black}{86.54} & \textcolor{blue!40!black}{78.26} & \cellcolor{blue!30}\textcolor{blue!40!black}{77.78} & \textcolor{blue!40!black}{75.95} & \cellcolor{blue!30}\textcolor{blue!40!black}{84.34} & \textcolor{blue!40!black}{81.25} & \textcolor{blue!40!black}{82.08} & \textcolor{blue!40!black}{63.64} & \textcolor{blue!40!black}{61.54} & \textcolor{blue!40!black}{77.78} & \cellcolor{blue!30}\textcolor{blue!40!black}{80.00} & \cellcolor{blue!30}\textcolor{blue!40!black}{90.48} & \textcolor{blue!40!black}{78.26} \\ \hline
        \textbf{Open-source} 
        & Qwen-VL& CoT & 52.66 & 43.66 & 59.14  & 55.00 & 35.71 & 57.69 & 56.52 & 22.22 & 44.30 & 62.65 & 25.00 & 51.89 & 36.36 & 23.08 & 33.33 & 62.50 & 47.62 & 60.87\\
        & Qwen-VL& EFL & \textcolor{blue!50!black}{65.00} & \textcolor{blue!50!black}{57.09} & \textcolor{blue!50!black}{70.70}  & \textcolor{blue!40!black}{70.00} & \textcolor{blue!40!black}{57.14} & \textcolor{blue!40!black}{73.08} & \textcolor{blue!40!black}{56.52} & \textcolor{blue!40!black}{44.44} & \textcolor{blue!40!black}{60.76} & \textcolor{blue!40!black}{75.90} & \textcolor{blue!40!black}{43.75} & \textcolor{blue!40!black}{65.09} & \textcolor{blue!40!black}{63.64} & \textcolor{blue!40!black}{38.46} & \textcolor{blue!40!black}{33.33} & \textcolor{blue!40!black}{70.00} & \textcolor{blue!40!black}{57.14} & \textcolor{blue!40!black}{65.22} \\
        & LLaVa-NEXT& CoT & 16.72 & 14.18 & 18.15 & 20.00 & 7.14 & 19.23 & 26.09 & 15.00 & 18.99 & 21.69 & 12.50 & 16.04 & 9.09 & 7.69 & 11.11 & 20.00 & 4.76 & 4.35\\
        & LLaVa-NEXT& EFL & \textcolor{blue!50!black}{28.28} & \textcolor{blue!50!black}{26.87} & \textcolor{blue!50!black}{29.30} & \textcolor{blue!40!black}{25.00} & \textcolor{blue!40!black}{17.14} & \textcolor{blue!40!black}{34.62} & \textcolor{blue!40!black}{26.09} & \textcolor{blue!40!black}{22.22} & \textcolor{blue!40!black}{29.11} & \textcolor{blue!40!black}{38.55} & \textcolor{blue!40!black}{12.50} & \textcolor{blue!40!black}{31.13} & \textcolor{blue!40!black}{27.27} & \textcolor{blue!40!black}{30.77} & \textcolor{blue!40!black}{11.11} & \textcolor{blue!40!black}{27.50} & \textcolor{blue!40!black}{14.29} & \textcolor{blue!40!black}{30.43} \\
        & LLaMa-3.2-Vision& CoT & 19.38 & 13.81 & 23.39  & 5.00 & 20.30 & 21.15 & 43.48 & 43.00 & 2.53 & 18.07 & 12.50 & 15.09 & 27.27 & 15.38 & 22.22 & 34.17 & 28.57 & 26.09 \\
        & LLaMa-3.2-Vision& EFL & \textcolor{blue!50!black}{27.19} & \textcolor{blue!50!black}{22.01} & \textcolor{blue!50!black}{30.91} & \textcolor{blue!40!black}{15.00} & \textcolor{blue!40!black}{29.20} & \textcolor{blue!40!black}{30.77} & \textcolor{blue!40!black}{47.83} & \textcolor{blue!40!black}{77.00} & \textcolor{blue!40!black}{10.13} & \textcolor{blue!40!black}{24.10} & \textcolor{blue!40!black}{31.25} & \textcolor{blue!40!black}{23.58} & \textcolor{blue!40!black}{36.36} & \textcolor{blue!40!black}{38.46} & \textcolor{blue!40!black}{33.33} & \textcolor{blue!40!black}{43.33} & \textcolor{blue!40!black}{38.10} & \textcolor{blue!40!black}{30.43} \\ \hline
        \end{tabular}
    }
\end{table}

We now present and analyze our experiment results in detail. Detailed results for mathematical reasoning and expertise reasoning are presented in Table \ref{tab:Evaluation Result 1} and Figure \ref{fig:teaser}(b), while performance across different financial topics are shown in Table \ref{tab:Evaluation Result 1} and Figure \ref{fig:teaser} (c). We summarize and analyze key findings as follows:

\noindent \textbf{Disparity between Open-source Models and Closed-source Models:} The results on \textbf{FinMR} reveal insights into the comparative performance of state-of-the-art LLMs and MLLMs, as highlighted in Table \ref{tab:Evaluation Result 1} (marked in blue and green). Closed-source models consistently outperform open-source counterparts. In particular, the textual LLM DeepSeek-R1 and multimodal Gemini-1.5-Pro gained 71.88\% and 82.06\% overall accuracy, respectively. In contrast, open-source models such as LLaMa 3.2 and LLaVa-NEXT demonstrate significantly lower overall performances, with accuracies falling below 30\%. These results highlight a critical disparity between open-source and closed-source models.
Open-source multimodal models like LLaVa\_NEXT and LLaMa\_3.2\_Vision performed below expectations, indicating the challenges faced by open-source approaches in achieving competitive performance on complex multimodal tasks.

\noindent \textbf{Effectiveness of Error Feedback Learning:} The comparison between CoT prompting and EFL highlights the effectiveness of leveraging error feedback to enhance model performance. Across all evaluated models,  accuracy improved significantly when EFL was applied. For example,  GPT\_o1, a reasoning-focused model, initially scored 46.56\% but saw a significant 13.75\% improvement (reaching 60.31\%) after incorporating  negative examples with constructive feedback. Similarly,  multimodal models Gemini-1.5-Pro and Qwen\_VL achieved nearly 12\% improvements after adopting EFL. This underscores the potential of high-quality error databases in refining reasoning capabilities. Moreover, the consistent performance gains across diverse models demonstrate that EFL is a robust and generalizable approach for improving reasoning in the financial domain.

\noindent \textbf{Impact of Image Captions and Direct Image Inputs:} As discussed above, for LLMs that lack inherent visual processing capabilities, images were captioned using GPT-4o to supplement their visual understanding. Textual-modality models trained on diverse datasets delivered moderate performance, with most achieving over 40\% accuracy, except for LLaMa 3.2, which struggled to perform on par with its peers. Among textual models, DeepSeek-R1 (EFL) stood out,  achieving 71.88\% accuracy and outperforming other closed-source LLMs such as GPT-o1. In contrast, MLLMs utilizing direct image input demonstrated significantly higher accuracy. Gemini-1.5-Pro (EFL) excelled as the top-performing multimodal model, achieving 82.06\% accuracy,  and slightly surpassing Claude-3.5-Sonnet. This highlights the enhanced reasoning capabilities enabled by direct image inputs. By enabling richer context through integrated textual and visual information, multimodal models demonstrate their superiority in addressing intricate, knowledge-intensive tasks. 

\noindent \textbf{Challenge of Financial Math Reasoning:} Table \ref{tab:Evaluation Result 1} reveals a significant gap between mathematical and expertise reasoning tasks. Models like GPT-o1, Claude-3.5-Sonnet, and LLaMA-3.2 (EFL) show lower accuracy in mathematical reasoning (e.g., 32.09\% for LLaMA-3.2) compared to expertise reasoning (38.98\%), due to the former's need for logical rigor and multi-step calculations. In contrast, expertise reasoning relies more on contextual understanding. Closed-source multimodal models like Gemini-1.5-Pro (EFL) achieve over 80\% accuracy in mathematical reasoning, outperforming expertise reasoning (83.27\% vs. 81.18\%). Figure \ref{fig:teaser}(b) confirms that multimodal inputs enhance performance in both tasks, though challenges remain in integrating textual and visual information. Visual results in Figure \ref{fig:teaser}(c) show that multimodal inputs significantly improve performance in numerically complex topics like Valuation and Risk Models (VRM) and Fixed Income (FI), but less so in less structured topics like Operational Risk and Economics. Although math reasoning tasks are more challenging, visual input makes a contribution to reasoning accuracy. Future work should focus on improving text-image interaction. 

\vspace{-0.3 cm}
\subsection{Error Type Analysis}
\vspace{-0.3 cm}
To gain deeper insights into the limitations of the tested models, we conducted an error analysis to identify common challenges encountered during the reasoning process. Errors were categorized into five primary types: \textbf{\textit{image recognition failure, question misunderstanding, incorrect formula application, answer not found}}, and \textbf{\textit{others}}. Figure~\ref{fig:teaser}(d) summarizes the prevalence of each error type, while Table~\ref{tab:error_analysis} provides representative examples.

Among these categories, \textbf{image recognition failures} emerged as the most significant one, accounting for 72.84\% of total errors. As shown in Table~\ref{tab:error_analysis}, many reasoning steps reveal that the provided images lack sufficient direct information for problem-solving. Moreover, a substantial portion of these images require domain-specific expertise to extract implicit information effectively, which current models struggle to achieve. This highlights the need for more sophisticated visual understanding capabilities, particularly in tasks involving specialized financial visuals such as charts and tables. 

Another prominent issue is \textbf{question misunderstanding}, particularly for questions within specialized financial domains at the college level. Models frequently misinterpret the intent or nuances of these questions, leading to incorrect reasoning steps. This limitation underscores the importance of integrating deeper contextual understanding into models, especially for domain-specific tasks. 

Even when questions and images are correctly interpreted, models often fail in \textbf{applying the correct formulas}. This issue is especially prevalent in harder-level questions requiring the integration of knowledge across multiple topics or the application of cross-domain financial formulas. These errors suggest that current models lack the ability to handle the logical rigor and multi-step calculations necessary for complex mathematical reasoning tasks. 

The \textbf{\textit{answer not found}} error type is another recurring issue, particularly for models like LLaMa-3.2-Vision and LLaVa-NEXT, which often fail to produce a final answer during the reasoning process. Our study revealed that for these models, unrestricted output tokens resulted in a \textbf{repetition problem}, where the reasoning output becomes repetitive. This issue, as shown in the first example of Table~\ref{tab:error_analysis}, leads to inconsistencies, overly lengthy reasoning steps, and ultimately incomplete answers. This phenomenon is particularly detrimental for tasks requiring long reasoning chains and highlights the importance of managing token limits.

In summary, these errors arise from technical limitations, such as image recognition failures and output repetition, and a lack of financial domain expertise. While our analysis provides a foundational understanding of error types, space limitations prevent us from presenting a full systematic error analysis. We believe that such an in-depth analysis would provide meaningful insights into the reasoning capabilities and shortcomings of large models and should be a focus of future work.
\vspace{-0.5 cm}
\section{Conclusion and Future Work}
\vspace{-0.3 cm}
This paper introduced \textbf{FinMR}, a new benchmark tailored to evaluate the financial reasoning capabilities of multimodal models. Through evaluations of open-source and closed-source LLMs and MLLMs, we identified key insights and highlighted critical challenges in this domain.
Our findings demonstrate three main conclusions: \textbf{(1)} MLLMs significantly outperform LLMs by effectively integrating textual and visual information, underscoring the importance of robust multimodal reasoning frameworks. However, image recognition remains a major bottleneck.
\textbf{(2)} The Error Feedback Learning (EFL) method consistently outperformed Chain of Thought (CoT) prompting, validating the efficacy of leveraging negative examples with feedback to improve reasoning. \textbf{(3)} Financial math reasoning tasks consistently pose greater difficulty, with models achieving approximately 10\% lower accuracy compared to expertise reasoning. Key challenges include incorrect formula application and question misunderstandings.
Future work should focus on improving models' visual reasoning abilities, exploring more efficient, training-free methods to enhance reasoning, and systematically addressing the challenges identified in this study to enable more robust financial reasoning systems. A more detailed and systematic analysis of errors arising during the reasoning process of large models would provide deeper insights into the specific limitations and failure modes of both LLMs and MLLMs, particularly in complex tasks like financial reasoning. 

\clearpage
\bibliography{FMR}

\begin{thebibliography}{45}
\providecommand{\natexlab}[1]{#1}
\providecommand{\url}[1]{\texttt{#1}}
\expandafter\ifx\csname urlstyle\endcsname\relax
  \providecommand{\doi}[1]{doi: #1}\else
  \providecommand{\doi}{doi: \begingroup \urlstyle{rm}\Url}\fi

\bibitem[AI@Meta(2024)]{ai@meta2024Llama32Revolutionizing}
AI@Meta.
\newblock Llama 3.2: {{Revolutionizing}} edge {{AI}} and vision with open, customizable models, 2024.
\newblock URL \url{https://ai.meta.com/blog/llama-3-2-connect-2024-vision-edge-mobile-devices/}.

\bibitem[Anthropic(2024)]{anthropicIntroducingComputerUse2024}
Anthropic.
\newblock Introducing computer use, a new claude 3.5 sonnet, and claude 3.5 haiku {\textbackslash} anthropic, 2024.
\newblock URL \url{https://www.anthropic.com/news/3-5-models-and-computer-use}.

\bibitem[Bai et~al.(2023)Bai, Bai, Yang, Wang, Tan, Wang, Lin, Zhou, and Zhou]{baiQwenVLVersatileVisionlanguage2023}
Jinze Bai, Shuai Bai, Shusheng Yang, Shijie Wang, Sinan Tan, Peng Wang, Junyang Lin, Chang Zhou, and Jingren Zhou.
\newblock Qwen-{{VL}}: {{A}} versatile vision-language model for understanding, localization, text reading, and beyond, 2023.
\newblock URL \url{http://arxiv.org/abs/2308.12966}.

\bibitem[Chen et~al.(2022{\natexlab{a}})Chen, Li, Qin, Lu, Lin, Chen, and Liang]{chen2022unigeounifyinggeometrylogical}
Jiaqi Chen, Tong Li, Jinghui Qin, Pan Lu, Liang Lin, Chongyu Chen, and Xiaodan Liang.
\newblock Unigeo: Unifying geometry logical reasoning via reformulating mathematical expression, 2022{\natexlab{a}}.
\newblock URL \url{https://arxiv.org/abs/2212.02746}.

\bibitem[Chen et~al.(2022{\natexlab{b}})Chen, Tang, Qin, Liang, Liu, Xing, and Lin]{chen2022geoqageometricquestionanswering}
Jiaqi Chen, Jianheng Tang, Jinghui Qin, Xiaodan Liang, Lingbo Liu, Eric~P. Xing, and Liang Lin.
\newblock Geoqa: A geometric question answering benchmark towards multimodal numerical reasoning, 2022{\natexlab{b}}.
\newblock URL \url{https://arxiv.org/abs/2105.14517}.

\bibitem[Chen et~al.(2024)Chen, Qin, Zhang, Chen, Xu, and Che]{chen2024M$^3$CoTnovelbenchmark}
Qiguang Chen, Libo Qin, Jin Zhang, Zhi Chen, Xiao Xu, and Wanxiang Che.
\newblock M$^3${CoT}: {A} novel benchmark for multi-domain multi-step multi-modal chain-of-thought.
\newblock arXiv, 2024.
\newblock URL \url{http://arxiv.org/abs/2405.16473}.

\bibitem[Chen et~al.(2023)Chen, Zhou, Shen, Hong, Zhang, and Gan]{chen2023seethinkconfirminteractive}
Zhenfang Chen, Qinhong Zhou, Yikang Shen, Yining Hong, Hao Zhang, and Chuang Gan.
\newblock See, think, confirm: Interactive prompting between vision and language models for knowledge-based visual reasoning, 2023.
\newblock URL \url{https://arxiv.org/abs/2301.05226}.

\bibitem[Chen et~al.(2021)Chen, Chen, Smiley, Shah, Borova, Langdon, Moussa, Beane, Huang, Routledge, and Wang]{chenFinQADatasetNumerical2021}
Zhiyu Chen, Wenhu Chen, Charese Smiley, Sameena Shah, Iana Borova, Dylan Langdon, Reema Moussa, Matt Beane, Ting-Hao Huang, Bryan Routledge, and William~Yang Wang.
\newblock {{FinQA}}: {{A}} dataset of numerical reasoning over financial data.
\newblock pp.\  3697--3711. Association for Computational Linguistics, 2021.
\newblock \doi{10.18653/v1/2021.emnlp-main.300}.
\newblock URL \url{https://aclanthology.org/2021.emnlp-main.300}.

\bibitem[Chen et~al.(2022{\natexlab{c}})Chen, Chen, Smiley, Shah, Borova, Langdon, Moussa, Beane, Huang, Routledge, and Wang]{chen2022FinQAdatasetnumerical}
Zhiyu Chen, Wenhu Chen, Charese Smiley, Sameena Shah, Iana Borova, Dylan Langdon, Reema Moussa, Matt Beane, Ting-Hao Huang, Bryan Routledge, and William~Yang Wang.
\newblock {{FinQA}}: {{A}} dataset of numerical reasoning over financial data.
\newblock arXiv, 2022{\natexlab{c}}.
\newblock URL \url{http://arxiv.org/abs/2109.00122}.

\bibitem[DeepSeek-AI et~al.(2025)DeepSeek-AI, Guo, and Yang]{deepseekai2025deepseekr1incentivizingreasoningcapability}
DeepSeek-AI, Daya Guo, and Dejian Yang.
\newblock Deepseek-r1: Incentivizing reasoning capability in llms via reinforcement learning, 2025.
\newblock URL \url{https://arxiv.org/abs/2501.12948}.

\bibitem[Dubey et~al.(2024)Dubey, Jauhri, Pandey, and Kadian]{dubeyLlama3Herd2024}
Abhimanyu Dubey, Abhinav Jauhri, Abhinav Pandey, and Kadian.
\newblock The llama 3 herd of models, 2024.
\newblock URL \url{http://arxiv.org/abs/2407.21783}.

\bibitem[GeminiTeam et~al.(2024)GeminiTeam, Clay, Kocisky, Perz, Yu, and Howard]{geminiteam2024Geminifamilyhighly}
GeminiTeam, Natalie Clay, Tomas Kocisky, Bartek Perz, Dian Yu, and Howard.
\newblock Gemini: {{A}} family of highly capable multimodal models, 2024.
\newblock URL \url{http://arxiv.org/abs/2312.11805}.

\bibitem[Jerven(2013)]{jerven2013poor}
Morten Jerven.
\newblock Poor numbers--how we are misled by african development statistics and what to do about it (uzuazo etemire).
\newblock \emph{VR{\"U} Verfassung und Recht in {\"U}bersee}, 46\penalty0 (3):\penalty0 336--340, 2013.
\newblock URL \url{http://www.jstor.org/stable/43239700}.

\bibitem[Kojima et~al.(2022)Kojima, Gu, Reid, Matsuo, and Iwasawa]{kojimaLargelanguagemodels}
Takeshi Kojima, Shixiang~(Shane) Gu, Machel Reid, Yutaka Matsuo, and Yusuke Iwasawa.
\newblock Large language models are zero-shot reasoners.
\newblock In S.~Koyejo, S.~Mohamed, A.~Agarwal, D.~Belgrave, K.~Cho, and A.~Oh (eds.), \emph{Advances in Neural Information Processing Systems}, volume~35, pp.\  22199--22213. Curran Associates, Inc., 2022.
\newblock URL \url{https://proceedings.neurips.cc/paper_files/paper/2022/file/8bb0d291acd4acf06ef112099c16f326-Paper-Conference.pdf}.

\bibitem[Li et~al.(2024{\natexlab{a}})Li, Zhang, Zhang, Guo, Li, Zhang, Liu, Yuan, and Li]{li2024LLaVANeXTStrongerLLMs}
Kaichen Li, Hao Zhang, Renrui Zhang, Dong Guo, Feng Li, Yuanhan Zhang, Ziwei Liu, Chun Yuan, and Bo~Li.
\newblock {{LLaVA-NeXT}}: {{Stronger LLMs}} supercharge multimodal capabilities in the wild, 2024{\natexlab{a}}.
\newblock URL \url{https://llava-vl.github.io/blog/2024-05-10-llava-next-stronger-llms/}.

\bibitem[Li et~al.(2024{\natexlab{b}})Li, Huang, and Zhang]{li2024surveymultimodalcomposite}
Suyan Li, Fuxiang Huang, and Lei Zhang.
\newblock A survey of multimodal composite editing and retrieval, 2024{\natexlab{b}}.
\newblock URL \url{http://arxiv.org/abs/2409.05405}.

\bibitem[Liang et~al.(2023)Liang, Yang, Zhang, and Zhang]{liang2023inproceedingsliangUniMathFoundationalMultimodal2023addresssingapore}
Zhenwen Liang, Tianyu Yang, Jipeng Zhang, and Xiangliang Zhang.
\newblock Unimath.
\newblock pp.\  7126--7133. Association for Computational Linguistics, 2023.
\newblock \doi{10.18653/v1/2023.emnlp-main.440}.
\newblock URL \url{https://aclanthology.org/2023.emnlp-main.440}.

\bibitem[Liu et~al.(2024)Liu, Lyu, Min, Wang, Su, and Wang]{liu2024Retrievalaugmentedmultimodalchainofthoughts}
Bingshuai Liu, Chenyang Lyu, Zijun Min, Zhanyu Wang, Jinsong Su, and Longyue Wang.
\newblock Retrieval-augmented multi-modal chain-of-thoughts reasoning for large language models, 2024.
\newblock URL \url{http://arxiv.org/abs/2312.01714}.

\bibitem[Lu et~al.(2023)Lu, Qiu, Yu, Welleck, and Chang]{lu2023surveydeeplearning}
Pan Lu, Liang Qiu, Wenhao Yu, Sean Welleck, and Kai-Wei Chang.
\newblock A survey of deep learning for mathematical reasoning.
\newblock arXiv, 2023.
\newblock \doi{10.48550/arXiv.2212.10535}.
\newblock URL \url{http://arxiv.org/abs/2212.10535}.

\bibitem[Lu et~al.(2024)Lu, Bansal, Xia, Liu, Li, Hajishirzi, Cheng, Chang, Galley, and Gao]{luMathVistaEvaluatingMathematical2024}
Pan Lu, Hritik Bansal, Tony Xia, Jiacheng Liu, Chunyuan Li, Hannaneh Hajishirzi, Hao Cheng, Kai-Wei Chang, Michel Galley, and Jianfeng Gao.
\newblock {{MathVista}}: {{Evaluating}} mathematical reasoning of foundation models in visual contexts, 2024.
\newblock URL \url{http://arxiv.org/abs/2310.02255}.

\bibitem[MacKenzie(2008)]{mackenzie2008engine}
Donald MacKenzie.
\newblock \emph{An engine, not a camera: How financial models shape markets}, volume~48.
\newblock Mit Press, 2008.
\newblock \doi{https://doi.org/10.1353/tech.2007.0154}.

\bibitem[OpenAI(2024{\natexlab{a}})]{openaiIntroducingGPT4oMore2024}
OpenAI.
\newblock Introducing {{GPT-4o}} and more tools to {{ChatGPT}} free users {\textbar} {{OpenAI}}, 2024{\natexlab{a}}.
\newblock URL \url{https://openai.com/index/gpt-4o-and-more-tools-to-chatgpt-free/}.

\bibitem[OpenAI(2024{\natexlab{b}})]{openaiIntroductionOpenAIO12024}
OpenAI.
\newblock Introduction {{OpenAI}} o1, 2024{\natexlab{b}}.
\newblock URL \url{https://openai.com/o1/}.

\bibitem[Qwen(2024)]{teamqwenQwen25LLMExtendingBoundary2024}
Team Qwen.
\newblock Qwen2.5-{{LLM}}: {{Extending}} the boundary of {{LLMs}}, 2024.
\newblock URL \url{http://qwenlm.github.io/blog/qwen2.5-llm/}.

\bibitem[Shao et~al.(2024)Shao, Wang, Zhu, Xu, Song, Bi, Zhang, Zhang, Li, Wu, and Guo]{Shao2024DeepSeekMathPushingLimitsMathematicalReasoningOpenLanguageModels}
Zhihong Shao, Peiyi Wang, Qihao Zhu, Runxin Xu, Junxiao Song, Xiao Bi, Haowei Zhang, Mingchuan Zhang, Y.~K. Li, Y.~Wu, and Daya Guo.
\newblock {{DeepSeekMath}}: {{Pushing}} the limits of mathematical reasoning in open language models, 2024.
\newblock URL \url{http://arxiv.org/abs/2402.03300}.

\bibitem[Sun et~al.(2024)Sun, An, Tian, Nan, Liu, Liu, Shah, and Chen]{sun2024reviewmultimodalexplainable}
Shilin Sun, Wenbin An, Feng Tian, Fang Nan, Qidong Liu, Jun Liu, Nazaraf Shah, and Ping Chen.
\newblock A review of multimodal explainable artificial intelligence: Past, present and future.
\newblock arXiv, 2024.
\newblock \doi{10.48550/arXiv.2412.14056}.
\newblock URL \url{http://arxiv.org/abs/2412.14056}.

\bibitem[Tan et~al.(2024)Tan, Wei, Sun, Gao, Li, Yu, Guo, and Li]{tan2024Retrievalmeetsreasoning}
Cheng Tan, Jingxuan Wei, Linzhuang Sun, Zhangyang Gao, Siyuan Li, Bihui Yu, Ruifeng Guo, and Stan~Z. Li.
\newblock Retrieval meets reasoning: {{Even}} high-school textbook knowledge benefits multimodal reasoning, 2024.
\newblock URL \url{http://arxiv.org/abs/2405.20834}.

\bibitem[Team et~al.(2024)Team, Georgiev, Lei, Burnell, Bai, and Gulati]{geminiGemini15Unlocking2024}
Gemini Team, Petko Georgiev, Ving~Ian Lei, Ryan Burnell, Libin Bai, and Gulati.
\newblock Gemini 1.5: {{Unlocking}} multimodal understanding across millions of tokens of context, 2024.
\newblock URL \url{http://arxiv.org/abs/2403.05530}.

\bibitem[Tsimpoukelli et~al.(2021)Tsimpoukelli, Menick, Cabi, Eslami, Vinyals, and Hill]{tsimpoukelliMultimodalFewshotLearning2021}
Maria Tsimpoukelli, Jacob Menick, Serkan Cabi, S.~M.~Ali Eslami, Oriol Vinyals, and Felix Hill.
\newblock Multimodal few-shot learning with frozen language models.
\newblock arXiv, 2021.
\newblock URL \url{http://arxiv.org/abs/2106.13884}.

\bibitem[Wang et~al.(2024{\natexlab{a}})Wang, Pan, Shi, Lu, Zhan, and Li]{wang2024Measuringmultimodalmathematical}
Ke~Wang, Junting Pan, Weikang Shi, Zimu Lu, Mingjie Zhan, and Hongsheng Li.
\newblock Measuring multimodal mathematical reasoning with {{MATH-vision}} dataset, 2024{\natexlab{a}}.
\newblock URL \url{http://arxiv.org/abs/2402.14804}.

\bibitem[Wang et~al.(2024{\natexlab{b}})Wang, Chen, Han, Lin, Zhao, Liu, Zhai, Yuan, You, and Yang]{wangExploringReasoningAbilities2024}
Yiqi Wang, Wentao Chen, Xiaotian Han, Xudong Lin, Haiteng Zhao, Yongfei Liu, Bohan Zhai, Jianbo Yuan, Quanzeng You, and Hongxia Yang.
\newblock Exploring the reasoning abilities of multimodal large language models ({{MLLMs}}): {{A}} comprehensive survey on emerging trends in multimodal reasoning, 2024{\natexlab{b}}.
\newblock URL \url{http://arxiv.org/abs/2401.06805}.

\bibitem[Wei et~al.(2022{\natexlab{a}})Wei, Tay, Bommasani, Raffel, Zoph, Borgeaud, Yogatama, Bosma, Zhou, Metzler, Chi, Hashimoto, Vinyals, Liang, Dean, and Fedus]{wei2022emergentabilitieslargelanguage}
Jason Wei, Yi~Tay, Rishi Bommasani, Colin Raffel, Barret Zoph, Sebastian Borgeaud, Dani Yogatama, Maarten Bosma, Denny Zhou, Donald Metzler, Ed~H. Chi, Tatsunori Hashimoto, Oriol Vinyals, Percy Liang, Jeff Dean, and William Fedus.
\newblock Emergent abilities of large language models, 2022{\natexlab{a}}.
\newblock URL \url{https://arxiv.org/abs/2206.07682}.

\bibitem[Wei et~al.(2022{\natexlab{b}})Wei, Wang, Schuurmans, Bosma, Ichter, Xia, Chi, Le, and Zhou]{wei2022Chainofthoughtpromptingelicits}
Jason Wei, Xuezhi Wang, Dale Schuurmans, Maarten Bosma, Brian Ichter, Fei Xia, Ed~H Chi, Quoc~V Le, and Denny Zhou.
\newblock Chain-of-thought prompting elicits reasoning in large language models.
\newblock 2022{\natexlab{b}}.
\newblock URL \url{http://arxiv.org/abs/2201.11903}.

\bibitem[Xue et~al.(2024)Xue, Chen, Zhou, Dai, Chu, and Mei]{xue2024fammabenchmarkfinancialdomain}
Siqiao Xue, Tingting Chen, Fan Zhou, Qingyang Dai, Zhixuan Chu, and Hongyuan Mei.
\newblock Famma: A benchmark for financial domain multilingual multimodal question answering, 2024.
\newblock URL \url{https://arxiv.org/abs/2410.04526}.

\bibitem[Yan et~al.(2024{\natexlab{a}})Yan, Geng, Cao, Li, Li, Li, Zhou, Yang, and Zhang]{yan2024TabMedBERTtabularknowledge}
Xu~Yan, Lei Geng, Ziqiang Cao, Juntao Li, Wenjie Li, Sujian Li, Xinjie Zhou, Yang Yang, and Jun Zhang.
\newblock {{TabMedBERT}}: {{A}} tabular knowledge enhanced biomedical pretrained language model.
\newblock IOS Press, 2024{\natexlab{a}}.
\newblock URL \url{https://ebooks.iospress.nl/doi/10.3233/FAIA240674}.

\bibitem[Yan et~al.(2024{\natexlab{b}})Yan, Su, He, Fu, Zheng, Lyu, Wang, Wang, Wen, and Hu]{yan2024surveymathematicalreasoning}
Yibo Yan, Jiamin Su, Jianxiang He, Fangteng Fu, Xu~Zheng, Yuanhuiyi Lyu, Kun Wang, Shen Wang, Qingsong Wen, and Xuming Hu.
\newblock A survey of mathematical reasoning in the era of multimodal large language model: Benchmark, method \& challenges, December 2024{\natexlab{b}}.
\newblock URL \url{http://arxiv.org/abs/2412.11936}.

\bibitem[Yu et~al.(2024)Yu, Tang, Xu, Cui, Ran, Yan, Liu, Wang, Han, Liu, and Sun]{yu2024VisRAGVisionbasedretrievalaugmented}
Shi Yu, Chaoyue Tang, Bokai Xu, Junbo Cui, Junhao Ran, Yukun Yan, Zhenghao Liu, Shuo Wang, Xu~Han, Zhiyuan Liu, and Maosong Sun.
\newblock {{VisRAG}}: {{Vision-based}} retrieval-augmented generation on multi-modality documents, 2024.
\newblock URL \url{http://arxiv.org/abs/2410.10594}.

\bibitem[Yue et~al.(2024)Yue, Ni, Zhang, Zheng, Liu, Zhang, Stevens, Jiang, Ren, Sun, Wei, Yu, Yuan, Sun, Yin, Zheng, Yang, Liu, Huang, Sun, Su, and Chen]{yueMMMUMassiveMultidiscipline2024}
Xiang Yue, Yuansheng Ni, Kai Zhang, Tianyu Zheng, Ruoqi Liu, Ge~Zhang, Samuel Stevens, Dongfu Jiang, Weiming Ren, Yuxuan Sun, Cong Wei, Botao Yu, Ruibin Yuan, Renliang Sun, Ming Yin, Boyuan Zheng, Zhenzhu Yang, Yibo Liu, Wenhao Huang, Huan Sun, Yu~Su, and Wenhu Chen.
\newblock {{MMMU}}: {{A}} massive multi-discipline multimodal understanding and reasoning benchmark for expert {{AGI}}, 2024.
\newblock URL \url{http://arxiv.org/abs/2311.16502}.

\bibitem[Zhang et~al.(2024)Zhang, Zhang, Li, Zhao, Karypis, and Smola]{zhang2024Multimodalchainofthoughtreasoning}
Zhuosheng Zhang, Aston Zhang, Mu~Li, Hai Zhao, George Karypis, and Alex Smola.
\newblock Multimodal chain-of-thought reasoning in language models, 2024.
\newblock URL \url{http://arxiv.org/abs/2302.00923}.

\bibitem[Zhao et~al.(2023{\natexlab{a}})Zhao, Chen, Wang, Jiao, Do, Qin, Ding, Guo, Li, Li, and Joty]{zhao2023Retrievingmultimodalinformation}
Ruochen Zhao, Hailin Chen, Weishi Wang, Fangkai Jiao, Xuan~Long Do, Chengwei Qin, Bosheng Ding, Xiaobao Guo, Minzhi Li, Xingxuan Li, and Shafiq Joty.
\newblock Retrieving multimodal information for augmented generation: {{A}} survey, 2023{\natexlab{a}}.
\newblock URL \url{http://arxiv.org/abs/2303.10868}.

\bibitem[Zhao et~al.(2022)Zhao, Li, Li, and Zhang]{zhao2022MultiHierttNumericalreasoning}
Yilun Zhao, Yunxiang Li, Chenying Li, and Rui Zhang.
\newblock {{MultiHiertt}}: {{Numerical}} reasoning over multi hierarchical tabular and textual data, 2022.
\newblock URL \url{http://arxiv.org/abs/2206.01347}.

\bibitem[Zhao et~al.(2023{\natexlab{b}})Zhao, Liu, Long, Zhang, Zhao, and Cohan]{zhao2023KnowledgeMathKnowledgeintensivemath}
Yilun Zhao, Hongjun Liu, Yitao Long, Rui Zhang, Chen Zhao, and Arman Cohan.
\newblock {{KnowledgeMath}}: {{Knowledge-intensive}} math word problem solving in finance domains, 2023{\natexlab{b}}.
\newblock URL \url{https://arxiv.org/abs/2311.09797}.

\bibitem[Zhao et~al.(2024{\natexlab{a}})Zhao, Liu, Long, Zhang, Zhao, and Cohan]{zhao2024FinanceMathKnowledgeintensivemath}
Yilun Zhao, Hongjun Liu, Yitao Long, Rui Zhang, Chen Zhao, and Arman Cohan.
\newblock {{FinanceMath}}: {{Knowledge-intensive}} math reasoning in finance domains, 2024{\natexlab{a}}.
\newblock URL \url{http://arxiv.org/abs/2311.09797}.

\bibitem[Zhao et~al.(2024{\natexlab{b}})Zhao, Long, Liu, Kamoi, Nan, Chen, Liu, Tang, Zhang, and Cohan]{zhao2024DocMathevalEvaluatingmath}
Yilun Zhao, Yitao Long, Hongjun Liu, Ryo Kamoi, Linyong Nan, Lyuhao Chen, Yixin Liu, Xiangru Tang, Rui Zhang, and Arman Cohan.
\newblock {{DocMath-eval}}: {{Evaluating}} math reasoning capabilities of {{LLMs}} in understanding long and specialized documents, 2024{\natexlab{b}}.
\newblock URL \url{http://arxiv.org/abs/2311.09805}.

\bibitem[Zhu et~al.(2021)Zhu, Lei, Huang, Wang, Zhang, Lv, Feng, and Chua]{zhu2021TATQAquestionanswering}
Fengbin Zhu, Wenqiang Lei, Youcheng Huang, Chao Wang, Shuo Zhang, Jiancheng Lv, Fuli Feng, and Tat-Seng Chua.
\newblock {{TAT-QA}}: {{A}} question answering benchmark on a hybrid of tabular and textual content in finance, 2021.
\newblock URL \url{http://arxiv.org/abs/2105.07624}.

\end{thebibliography}
\bibliographystyle{iclr2025_conference}

\clearpage
\appendix

\section{Appendix}
\subsection{Comparisons with Existing Benchmarks}
\label{A1: Comparisons with Existing Benchmarks}
We outline the differences between \textbf{FinMR} and 9 reasoning benchmarks: MathVista \citep{luMathVistaEvaluatingMathematical2024}, MMMU \citep{yueMMMUMassiveMultidiscipline2024}, MATH-V \citep{wang2024Measuringmultimodalmathematical}, FinQA \citep{chen2022FinQAdatasetnumerical}, TAT-QA \citep{zhu2021TATQAquestionanswering}, MultiHiertt \citep{zhao2022MultiHierttNumericalreasoning}, DocMath-Eval \citep{zhao2024DocMathevalEvaluatingmath}, FinanceMath \citep{zhao2024FinanceMathKnowledgeintensivemath}, and FAMMA \citep{xue2024fammabenchmarkfinancialdomain}. Table \ref{tab:comparison_benchmarks} provides a detailed comparison.

\noindent \textbf{Comparison with Multimodal Benchmarks.}
Math Vista \citep{luMathVistaEvaluatingMathematical2024}  is a consolidated mathematical reasoning benchmark in visual contexts, containing 6,141 examples categorized into seven reasoning types: \textit{algebraic reasoning, arithmetic reasoning, geometry reasoning, logical reasoning, numeric common sense, scientific reasoning,} and \textit{statistical reasoning}. While valuable, these tasks are primarily designed for elementary and high school levels.  MMMU \citep{yueMMMUMassiveMultidiscipline2024}, with 11500 examples, extends reasoning benchmarks to  college-level  tasks. It is designed to evaluate the multi-disciplinary multimodal understanding and reasoning capabilities of MLLMs, covering 30 subjects across six disciplines. MATH-V \citep{wang2024Measuringmultimodalmathematical} involves 2,252  challenging questions derived from diverse competition datasets, including Math Kangaroo, UK [Grey, Pink, Junior, Senior], ACM, and AIME. These questions emphasize expert-level visual perception and deliberate reasoning across 16 subjects.

Unlike Math Vista, MU, and MATH-V, \textbf{FinMR} is designed specifically to focus on financial reasoning. It combines domain-specific expertise with mathematical reasoning.  Furthermore, \textbf{FinMR} provides detailed, manually verified explanations for all QA pairs, averaging 61.43 words per explanation. 
As shown in Table \ref{tab:Benchmark Statistics}, our benchmark has an average question length of 33.97 words, alongside extra context (average 327.74 words), which significantly exceeds the averages of Math Vista (15.6 words), MU (59.33 words), and MATH-V (42.3 words). By offering such comprehensive content,\textbf{FinMR} supports more advanced reasoning tasks and ensures that models are challenged with realistic financial scenarios.

\noindent \textbf{Comparison with Financial QA Benchmarks.}
As summarized in Table~\ref{tab:comparison_benchmarks}, existing financial QA benchmarks primarily target LLMs and focus on specific subdomains within finance, with limited multimodal content. The sole exception, FAMMA \citep{xue2024fammabenchmarkfinancialdomain}, includes 1,758 QA pairs across eight topics related to the CFA exam. However, it omits key financial areas such as risk management, constraining its ability to comprehensively evaluate MLLMs. In contrast, \textbf{FinMR} incorporates 15 topics derived from CFA and FRM exams, offering broader coverage of essential financial concepts and enabling more thorough assessments of reasoning capabilities in MLLMs.

Benchmarks like FinQA \citep{chen2022FinQAdatasetnumerical}, TAT-QA \citep{zhu2021TATQAquestionanswering}, and MultiHiertt \citep{zhao2022MultiHierttNumericalreasoning} primarily focus on numerical reasoning over financial tables extracted from real-world reports, such as earnings statements and financial accounting documents. While these datasets include large numbers of QA pairs: 8,281, 13,215, and 10,440 examples, respectively, they emphasize simpler numerical calculations or specific subtopics like financial reporting without addressing broader or more complex financial reasoning tasks. In contrast, \textbf{FinMR} includes mathematical reasoning questions that require advanced knowledge in mathematics and statistics, such as calculus, and spans seven distinct image types, including complex financial tables, as illustrated in Figure~\ref{fig:teaser}(a).

Regarding the format of reasoning steps, some benchmarks such as DocMath-Eval \citep{zhao2024DocMathevalEvaluatingmath} and FinanceMath \citep{zhao2024FinanceMathKnowledgeintensivemath} utilize Python-based solutions for interpretability. While these code-based explanations aim to provide precision, they often lack intuitive readability, making error analysis more challenging. \textbf{FinMR} addresses this limitation by including detailed, manually annotated textual explanations for all 3,200 QA pairs. These explanations offer a richer and more interpretable resource for analyzing reasoning steps, facilitating a deeper understanding of models' strengths and limitations.

\clearpage
\subsection{Error Type Analysis}
\label{A3: Error Type Analysis}
\begin{table}[h]
    \centering
    \caption{Model Reasoning Error Analysis}
    \label{tab:error_analysis}
    \resizebox{\columnwidth}{!}{ 
    \begin{tabular}{p{2cm} p{2cm} p{4cm} p{4cm}}
        \toprule
        \textbf{Error Type} & \textbf{Model} & \textbf{Model Reasoning Steps} & \textbf{Human Check / Explanation} \\
        \midrule

        Answer Not Found: Repetition Problem & LLaMa-3.2-Vision & 
        \textcolor{red}{
        Now, let's calculate the present value of the face amount at maturity:}
        \( PV = \frac{\$100}{(1+0.03)^7} \approx \$64.91 \)
        ...
        \textcolor{red}{Now, let's calculate the present value of the face amount at maturity:}
        
        & The reasoning step repetition results in no final answer. \\

        \midrule
        Wrong Financial Math Formula & LLaVa\_NEXT; Qwen2\_VL & 
        ..... The distance to default:
        \( \text{DD} = \frac{\ln\left(\frac{V}{D}\right) + \left(r - \frac{\sigma^2}{2}\right)T}{\sigma \sqrt{T}} \)
        & The distance to default :
        \(
        \textcolor{red}{\text{DD} = \frac{\text{Asset value} - \text{Default Point}}{\text{Asset volatility}}}
        \)
        \\

        \midrule
        Question Misunderstanding & LLaVa\_NEXT & 
        This appears to be a task related to logic puzzles, \textcolor{red}{specifically an example of a "river crossing" problem} where you need to ...
        & No "river crossing" problem in this case. \\

        \midrule
        Image Recognition Problem & LLaVa\_NEXT; Qwen2\_VL & 
        ... \textcolor{red}{However, the problem does not provide the values of Q1 and Q3. Without these specific values,} ...
        & This image is inputted; the model should recognize the values instead of assuming their absence. \\

        \midrule
        Image Recognition Problem & GPT-4o & 
        \textcolor{red}{.... based on the graph alone, the spot rate should be understood as 4.0\%.}

        &
        \(
        r(5) = 5 \sqrt{\frac{1.0437}{0.8394}} - 1 = 4.453\%
        \)
        \textcolor{red}{The model needs to extract data from the image for calculation instead of relying solely on textual information.} \\

        \bottomrule
    \end{tabular}
    }
\end{table}

\end{document}